\documentclass[journal,oneside]{IEEEtran}

\usepackage{amsfonts, amsmath, amssymb, amsthm}
\usepackage{algorithm}
\usepackage{url}
\usepackage{algpseudocode}
\usepackage{verbatim}
\usepackage{multirow}
\usepackage{graphicx}
\usepackage{hhline}
\usepackage{caption}
\usepackage{subcaption}
\usepackage{placeins}
\usepackage[table]{xcolor}
\MakeRobust{\Call}

\newcommand{\R}{\mathbb{R}}

\newcommand{\E}{\mathbb{E}}

\graphicspath{{../figures/}}

\begin{document}
\title{Tensor Denoising via Amplification and Stable Rank Methods}

\author{Jonathan Gryak$^{1}$, Kayvan Najarian$^{2,3,4,5,6}$, and Harm Derksen$^{7}$
\thanks{$^{1}$Department of Computer Science, Queens College, City University of New York, New York, NY, USA}%
\thanks{$^{2}$Department of Computational Medicine and Bioinformatics, University of Michigan, Ann Arbor, MI, USA}%
\thanks{$^{3}$Department of Emergency Medicine, University of Michigan, Ann Arbor, MI, USA}%
\thanks{$^{4}$Electrical and Computer Engineering, College of Engineering, University of Michigan, Ann Arbor, MI, USA}%
\thanks{$^{5}$Michigan Institute for Data Science, University of Michigan, Ann Arbor, MI, USA}%
\thanks{$^{6}$Max Harry Weil Institute for Critical Care Research and Innovation, University of Michigan, Ann Arbor, MI, USA}%
\thanks{$^{7}$Department of Mathematics, Northeastern University, Boston, MA, USA}
}

\maketitle

\begin{abstract}
Tensors in the form of multilinear arrays are ubiquitous in data science applications. Captured real-world data, including video, hyperspectral images, and discretized physical systems, naturally occur as tensors and often come with attendant noise. Under the additive noise model and with the assumption that the underlying clean tensor has low rank, many denoising methods have been created that utilize tensor decomposition to effect denoising through low rank tensor approximation. However, all such decomposition methods require estimating the tensor rank, or related measures such as the tensor spectral and nuclear norms, all of which are NP-hard problems.\\

In this work we leverage our previously developed framework of \textit{tensor amplification}, which provides good approximations of the spectral and nuclear tensor norms, to denoising synthetic tensors of various sizes, ranks, and noise levels, along with real-world tensors derived from physiological signals. We also introduce two new notions of tensor rank -- \textit{stable slice rank} and \textit{stable $X$-rank} -- and new denoising methods based on their estimation. The experimental results show that in the low rank context, tensor-based amplification provides comparable denoising performance in high signal-to-noise ratio (SNR) settings and superior performance in noisy (i.e., low SNR) settings, while the stable $X$-rank method achieves superior denoising performance on the physiological signal data.
\end{abstract}


\IEEEpeerreviewmaketitle

\section{Introduction}
\IEEEPARstart{T}{ensors} in the form of multilinear arrays are ubiquitous in data science applications. Captured real-world data, including color and hyperspectral images (HSIs), video, and discretized physical systems, naturally occur as tensors and often come with attendant noise. As is common in other signal processing applications, the captured tensor $\mathcal{T}\in \R^{p_1\times p_2 \times \cdots \times p_d}$ is modeled as $\mathcal{T}=\mathcal{D}+\mathcal{N}$, where $\mathcal{D}$ is a pure or ``clean" tensor $\mathcal{D}$ that has been corrupted by additive noise $\mathcal{D}$, which is typically assumed to be Gaussian. Additionally, the clean tensor $\mathcal{D}$ is assumed to be low rank.

Under this framework, tensor denoising can be achieved by utilizing tensor decompositions methods, such as the canonical polyadic (CP) \cite{carroll1970analysis,harshman1970foundations} and Tucker \cite{tucker1963implications,tucker1964extension} decompositions, to determine a low-rank approximation of the observed tensor. These decomposition algorithms require a pre-specified rank to compute an approximation, however, determining the rank of a tensor is NP-hard \cite{hillar2013most}. Thus, tensor decomposition-based methods utilize some estimate of the tensor rank to effect tensor denoising.

CP decomposition has been frequently used for HSI denoising, such as in Liu et al., \cite{liu2012denoising}, which estimated the tensor rank using covariance matrices of the $n$-model flattenings; in Veganzones et al. \cite{veganzones2015nonnegative}, which used a non-negative variant of CP decomposition; and in \cite{xue2019nonlocal}, in which a CP decomposition regularized by the nuclear norm of clustered 3D patches of the HSI was employed. Tucker decomposition based denoising include two works by Rajwade et al. that utilized higher order singular value decomposition \cite{de2000multilinear}, a tensor analog of matrix SVD, to denoise video \cite{rajwade2011using} and images \cite{rajwade2012image}; as well Lee et al. \cite{lee2020tensor}, which focused on denoising tensors with ordinal values. More recently, a tensor train (matrix product state) decomposition \cite{oseledets2011tensor} was used for denoising of HSIs \cite{gong2020tensor}.

A general framework for understanding tensor denoising in the additive model was developed in \cite{derksen2018general}, that relates the problem of denoising in the low rank context to the minimization of dual norms $\Vert \mathcal{D} \Vert_X$ and $\Vert \mathcal{N} \Vert_Y$, such as the nuclear $\Vert \mathcal \cdot \Vert_{\star}$ and spectral $\Vert \mathcal \cdot \Vert_{\sigma}$ norms, respectively. The calculation of these norms for tensors is also NP-hard \cite{friedland2018nuclear},\cite{hillar2013most}, thus in order make use of the denoising framework in \cite{derksen2018general} the co-authors developed the method of tensor amplification \cite{tokcan2021algebraic}, which provides good approximations of the tensor spectral norm and its dual the nuclear norm.

In this work we build upon the general denoising framework introduced in \cite{derksen2018general} to devise three novel tensor denoising methods. The first method uses \textit{tensor amplification} as previously developed by the co-authors in  \cite{tokcan2021algebraic}. The other two methods utilize estimates of new notions of tensor rank, \textit{stable slice rank} and \textit{stable $X$-rank}, which are introduced in Section \ref{sec:newmethods} and can be considered ``stable" in the sense that they are robust to perturbations of a tensor's values. 
The performances of these three new denoising algorithms are compared to several standard decomposition-based denoising methods -- CP Alternating Least Squares (CP-ALS) \cite{tensortoolbox}, a multiway (multilinear) Wiener filter \cite{muti2005multidimensional}, and higher order orthogonal iteration (HOOI) \cite{de2000best} -- on synthetic tensors of various sizes, ranks, and noise levels. These methods are also evaluated on real-world tensors derived from electrocardiogram (ECG) signals collected from healthy subjects and patients with various cardiac arrhythmias. The experimental results show that in the low rank context, tensor-based amplification provides comparable denoising performance in high signal-to-noise ratio (SNR) settings and superior performance in noisy (i.e., low SNR) settings, while the stable $X$-rank method achieves superior denoising performance on the ECG data.
\section{Preliminaries and Related Work}
\subsection{Basic Notation}
\label{sec:notation}
Let $\mathcal{T}\in \R^{p_1\times p_2 \times \cdots \times p_d}$ denote a real-valued tensor of  order $d$. In the denoising experiments that are performed in this study we will assume that the tensor $\mathcal{T}$ is the noisy version of a pure tensor $\mathcal{D}\in \R^{p_1\times p_2 \times \cdots \times p_d}$ corrupted by additive noise $\mathcal{N}\in \R^{p_1\times p_2 \times \cdots \times p_d}$, that is,
\begin{equation}
\mathcal{T}=\mathcal{D}+\mathcal{N}.
\end{equation}

The \emph{Frobenius norm} of $\mathcal{T}$ is denoted $\Vert\mathcal{T}\Vert$ and defined as 
\begin{equation}
\Vert\mathcal{T}\Vert=\sqrt{\sum_{i_1}^{p_1}\sum_{i_2}^{p_2}\cdots\sum_{i_d}^{p_d}t_{i_1i_2\cdots i_d}^2},
\end{equation}
while the \emph{tensor inner product} of two tensors $\mathcal{T},\mathcal{S}$ of matching order and dimension is defined as
\begin{equation}
\langle\mathcal{T},\mathcal{S}\rangle=\sum_{i_1}^{p_1}\sum_{i_2}^{p_2}\cdots\sum_{i_d}^{p_d}t_{i_1i_2\cdots i_d}s_{i_1i_2\cdots i_d}.
\end{equation}
The induced norm of the tensor inner product is the Frobenius norm defined above, with the typical relation $\langle\mathcal{T},\mathcal{T}\rangle=\Vert\mathcal{T}\Vert^2$.

\

Given a tensor $\mathcal{T}$ and a permutation $\mathbf{q}=\langle q_1,\ldots,q_d\rangle$ of the indices $1:d$, the $\mathbf{q}$\textit{-transpose} of $\mathcal{T}$ is the tensor $\mathcal{T}^{\langle \mathbf{q}\rangle}\in\R^{p_{q_1}\times p_{q_2} \times \cdots \times p_{q_d}}$ with entries
\begin{equation}
(\mathcal{T}^{\langle \mathbf{q}\rangle})_{i_1i_2\ldots i_d}=t_{i_{q_1}i_{q_2}\ldots i_{q_d}}. 
\end{equation}

At times we will need to \emph{matricize} the tensors under consideration by rearranging their entries in specific ways, as well as employ various tensor-tensor, tensor-matrix, and matrix-matrix products. In the definitions below and throughout the manuscript we will primarily follow the notational conventions introduced by Kolda and Bader in \cite{kolda2009tensor}.

\

The \emph{mode-$n$ flattening} or unfolding of the tensor $\mathcal{T}$ is the matrix $T_{(n)}\in \R^{p_n\times N/{p_n}}$, where $N=\prod_i p_i$, whose columns are the \textit{mode-$n$ fibers} of $\mathcal{T}$.
\

The \emph{$n$-mode product} of a tensor $\mathcal{T}$ and a matrix $A\in\R^{J\times p_n}$ is the tensor $\mathcal{T}\times_n A$ of size $p_1\times p_2 \times \cdots \times p_{n-1} \times J \times p_{n+1} \times \cdots \times p_d$ with entries
\begin{equation}
    (\mathcal{T}\times_n A)_{i_1i_2\ldots i_{n-1}j i_{n+1}\ldots i_d} = \sum_{i_n=1}^{p_n}t_{i_1i_2\ldots i_d}u_{j i_n}.
\end{equation}
If $\mathcal{S}=\mathcal{T}\times_n A$, then the $n$-mode product as defined above is equivalent to $S_{(n)}=AT_{(n)}$.
\

The \emph{Kronecker product} of two matrices $A\in\R^{I\times J}$ and $B\in\R^{K\times L}$ is the matrix $A\otimes B\in\R^{IK\times JL}$ defined by
\begin{equation}
    A\otimes B=\left[
    \begin{array}{cccc}
    a_{11}B&a_{12}B&\cdots&a_{1J}B\\
    a_{21}B&a_{22}B&\cdots&a_{2J}B\\
    \vdots&\vdots&\ddots&\vdots\\
    a_{I1}B&a_{I2}B&\cdots&a_{IJ}B
    \end{array}\right].
\end{equation}
\

Finally, given two tensors $\mathcal{T}\in\R^{p_1\times\cdots\times p_d}$ and $\mathcal{S}\in\R^{q_1\times\cdots\times q_e}$, their \emph{outer product} is the tensor $\mathcal{T}\circ\mathcal{S}$ of size $p_1\times\cdots\times p_d\times q_1\times\cdots\times q_e$ with entries
\begin{equation}
    (\mathcal{T}\circ \mathcal{S})_{i_1i_2\ldots i_d j_1j_2\ldots j_e} = t_{i_1} t_{i_2}\ldots t_{i_d} s_{j_1} s_{j_2}\ldots s_{j_e}. 
\end{equation}
\subsection{Decomposition-based Denoising}
\label{sec:decompdenoising}
Tensor decomposition methods seek to represent a given tensor by decomposing it into factors such as simple tensors or matrices and whose combination results in a ``good" approximation of the original tensor. In the context of denoising, it is typical to assume that a noisy signal is sparse, in the sense that its $\ell_1$ norm is small. In the case of matrices and tensors, this assumption corresponds to the original tensor having low rank, with the high rank components corresponding to additive noise. Thus, computing a low rank approximation of the original tensor via tensor decomposition is a means to effect tensor denoising.

In the case of matrices (order two tensors), singular value decomposition yields the exact rank $r$ of the matrix and its decomposition into $r$ factor, with the best low rank approximation for a given rank $l<r$ provided by choosing the factors corresponding to the $l$ largest singular values. For higher order tensors, calculating the exact rank is NP hard \cite{hillar2013most}. Moreover, unlike the matrix case, the factors used to create the best rank $r-1$ approximation need not be those used to produce the best rank $r$ approximation \cite{kolda2003counterexample}, and for \textit{degenerate} tensors, the best rank $r$ approximation may not even exist \cite{de2008tensor}.

Despite these theoretical limitations, in practice one can utilize tensor decomposition methods to effect denoising by creating decompositions for a range of rank values, then choosing the best rank $r$ decomposition $\mathcal{D}$ that best approximates the original tensor $\mathcal{T}$, e.g., $\min \Vert \mathcal{T}-\mathcal{D}\Vert$. This strategy for tensor denoising was evaluated using three common tensor decomposition methods: \textit{canonical polyadic decomposition}, \textit{higher-order orthogonal iteration}, and \textit{multiway Wiener filters}.

\

\subsubsection{CP Decomposition via Alternating Least Squares (CP-ALS)}
Let $U^{(j)}=[u_{j,1}u_{j,2}\ldots u_{j,r}] \in \mathbb{R}^{p_j \times r},~1\leq j \leq d.$  \emph{CP decomposition} factorizes a $d$-way tensor into $d$ factor matrices and a vector $\Lambda=[\lambda_1, \lambda_2,\ldots, \lambda_r] \in \mathbb{R}^{r}$:
\begin{equation}
    \mathcal{S}=\sum_{i=1}^{r} \lambda_i u_{1,i}\circ u_{2,i}\circ \ldots \circ u_{d,i}.
\end{equation}
The best rank $r$ approximation problem for a  tensor $\mathcal{T}  \in \mathbb{R}^{p_1 \times p_2 \times \ldots \times p_d}$ can be given as:
\begin{equation}
\min_{ \Lambda, U^{(1)}, \ldots, U^{(d)}} \|{\mathcal T - \mathcal{S}}\|~\text{where}~\mathcal{S}=[\Lambda~;~U^{(1)},~U^{(2)},\ldots,~U^{(d)}].
\end{equation}
This can be found by employing \emph{alternating least squares (ALS)}, wherein each iteration of the algorithm an approximation of the flattening for one mode is found by fixing all other modes of the tensors and solving a least squares problem. This process is repeated, cycling through all modes, until convergence or a maximum number of iterations is reached. In this work, the implementation of CP-ALS from TensorToolbox \cite{tensortoolbox} was utilized with the default level of tolerance ($10^{-4}$) and maximum number of iterations (50). CP-ALS was run for specified rank values $r\in[1,\min(p_i)]$, with the rank $r^*$ approximation
\begin{equation}
\label{eqn:bestdenoised}
\mathcal{D}_{r^*}=\min_{r} \Vert \mathcal{T}-\mathcal{D}_{r}\Vert
\end{equation}
chosen as the best denoised tensor.
\subsubsection{Higher-Order Orthogonal Iteration (HOOI)}
For matrices, \emph{orthogonal iteration} produces a sequence orthonormal bases for each subspace of the vector space. De Lathauwer et al. \cite{de2000best} extended this to tensors, developing the technique known as \emph{higher-order orthogonal iteration} (HOOI). This method uses ALS to estimate the best $rank\textrm{-}[r_1,\ldots,r_d]$ approximation for a tensor, and is achieved by iteratively solving the optimization problem
\begin{equation}
\textrm{argmin}_{U_{(i)\mid r_i}} \Vert \mathcal{T}-\mathcal{G}\times_1U_{(1)\mid r_1}\times_2U_{(2)\mid r_2}\times\ldots\times_NU_{(d)\mid r_d} \Vert,
\end{equation}
where $\mathcal{G}$ is a core tensor of size $r_1 \times \ldots \times r_d$  and each $U_{(i)\mid r_i}$ is a matrix comprised of the $r_i$ leftmost singular vectors of the singular value decomposition of the modal flattening $U_{(i)}$.

HOOI-based denoising was implemented using the \texttt{tucker\_als} method in TensorToolbox \cite{tensortoolbox} to determine the best rank $[r_1^*,\ldots,r_d^*]$ approximation, where each $r_i$ was chosen equally and uniformly from $r\in[1,\min(p_i)]$, with the rank $r^*$ approximation (Eq. \ref{eqn:bestdenoised}) chosen as the best denoised tensor.
\subsubsection{Multiway Wiener Filter}
For a discrete signal $y[n]$ and filter output $\hat{y}[n]$, the \emph{Wiener filter} $h[n]$ is the filter that minimizes the mean squared error between $\hat{y}[n]$ and $y[n]$:
\begin{equation}
\textrm{argmin}_{h[\cdot]} \E[ (\hat{y}[n]-y[n])^2)].
\end{equation}
Wiener filters have been used in a variety of denoising applications, such as for images \cite{jin2003adaptive,zhang2016image}, physiological signals \cite{smital2012adaptive,somers2018generic} and speech \cite{spriet2004spatially,chen2006new}.

\

Muti et al. \cite{muti2005multidimensional}  created a \emph{multiway Wiener filter} that can be used to denoise tensors of arbitrary size. Given a noisy tensor $\mathcal{T}$, their method uses an ALS approach to learn Wiener filters $\{H_{n}\}$ for each mode $n$ so that the mean squared error between $\mathcal{T}$ and the denoised tensor $D$ is minimized, where
\begin{equation}
\mathcal{D}=\mathcal{T}\times_1H_1\times_2H_2\times_3\cdots\times_dH_d.
\end{equation}
The implementation of the multiway Wiener filter utilized in this study and the exposition below follows \cite{lin2013survey}. The filters $H_n$ in each mode $n$ are initialized to the identity matrix of $\R^{p_n}$. At each stage $k$ of the algorithm, the filter $H_n^k$ is computed for each mode as
\begin{equation}
H_n^k=V_{n}\Lambda_{n}V_{n}^{{\intercal}},
\end{equation}
where $V_{n}$ is a matrix containing the $K_n$ orthonormal basis vectors of the signal subspace in the column space of $T_{(n)}$, the mode-$n$ flattening of $\mathcal{T}$, and 
\begin{equation}
\label{Biglambda}
\Lambda_{n}=\mathrm{diag}\left(\frac{\lambda_1^{\gamma}-\hat{\sigma}^{\gamma^2}_{n}}{\lambda_1^{\Gamma}},\ldots,\frac{\lambda_{K_n}^{\gamma}-\hat{\sigma}^{\gamma^2}_{n}}{\lambda_{K_n}^{\Gamma}}\right),
\end{equation}
where $\{\lambda_i^{\gamma}, i=1,\ldots,K_n\}$ and $\{\lambda_i^{\Gamma}, i=1,\ldots,K_n\}$ are respectively the $K_n$ largest eigenvalues of the matrices $\gamma_{n}$ and $\Gamma_{n}$, defined as
\begin{eqnarray}
    \gamma_{n} =& \E\left[T_{(n)} q_{n} {T_{(n)}}^{\intercal}\right]\\ 
    \Gamma_{n} =& \E\left[T_{(n)} Q_{n} {T_{(n)}}^{\intercal}\right] 
\end{eqnarray}
with
\begin{eqnarray}
    q_{n} &=& \bigotimes_{i\neq n}^d H_{i}\\
    Q_{n} &=& \bigotimes_{i\neq n}^d H_{i}^{\intercal}H_{i}.
\end{eqnarray}
The values $\hat{\sigma}^{\gamma^2}_{n}$ in Equation \ref{Biglambda} are estimates of the $p_n-K_n$ smallest eigenvalues of $\gamma_{n}$, calculated as
\begin{equation}
    \hat{\sigma}^{\gamma^2}_{n}=\frac{1}{p_n-K_n}\sum_{i=K_n+1}^{p_n}\lambda_i^{\gamma}.
\end{equation}
Following \cite{renard2008denoising}, the optimal $K_n$ for mode $n$ is estimated using the Akaike Information Criterion (AIC). Please refer to  \cite{lin2013survey} for further details.

\section{Amplification and Stable Rank Denoising}
\label{sec:newmethods}
 In this section we introduce three novel denoising methods -- \textit{Amplification}-based, \textit{Stable Slice Rank}, and \textit{Stable $X$-Rank} denoising. Amplification-based denoising builds upon the dual norms framework introduced in \cite{derksen2018general} and the tensor spectral norm approximation developed in \cite{tokcan2021algebraic}, to effect tensor denoising using \textit{solely tensor-based operations} - no matricization is required. The other two methods are built around newly introduced methods of tensor rank that provide robust estimates on noisy data and overcome the potential for tensor degeneracy in iterative approximation methods that rely on tensor rank, such as the established methods introduced in Section \ref{sec:decompdenoising}.

\subsection{A Framework for Denoising Using Dual Norms}
The model $\mathcal{T}=\mathcal{D}+\mathcal{N}$ utilized in this work can be viewed as an instance of the additive noise model $c=a+b$, where $a,b,c\in V$ are elements of a vector space $V$. In Derksen \cite{derksen2018general}, a general framework for understanding the denoising of vectors under the additive model was developed that relates the problem of denoising the vector $c$ to the minimization of $\Vert a\Vert_X$ and $\Vert b\Vert_Y$, where $\Vert \cdot \Vert_X$ and $\Vert \cdot \Vert_Y$ are dual norms. Moreover, the framework makes the assumptions that the original vector (or tensor) $a$ is \textit{sparse}, e.g., that it has few non-zero values or is of low rank, while the additive noise $b$ is \textit{dense} or of high rank. Thus, the norms $\Vert\cdot\Vert_X$ and $\Vert\cdot\Vert_Y$ can be interpreted as respectively measuring the sparsity and noise of the vector (or tensor) under consideration. The prototypical $\Vert\cdot\Vert_X$ norm is the \textit{nuclear norm}, which for a matrix is the sum of its singular values, while for a tensor the \textit{tensor nuclear norm} $\Vert\mathcal{T}\Vert_{\star}$, is defined as
$$
\Vert\mathcal{T}\Vert_{\star}=\min \sum_{i=1}^r \Vert v_i\Vert_2,
$$
where $v_1,\ldots,v_r$ are rank-1 tensors and $\mathcal{T}=\sum_{i=1}^r v_i$.

The prototypical $\Vert\cdot\Vert_Y$ norm and dual to $\Vert\cdot\Vert_X$ is the \textit{spectral norm}, which for a matrix is the absolute value of its largest singular value, while for a tensor the \textit{tensor spectral norm} $\Vert\mathcal{T}\Vert_{\sigma}$ is defined as
$$
\Vert\mathcal{T}\Vert_{\sigma}=  \sup  | {\mathcal T}\cdot u_{1} \otimes u_{2} \otimes \ldots \otimes u_{d}|,
$$
where $u_{j} \in \mathbb{R}^{p_j}$ and $\|u_{j}\|=1$ for $1\leq j\leq d$.

If $V$ is also an inner product space we also have the induced norm $\sqrt{\langle c,c\rangle}$ that corresponds to the standard Euclidean norm $\Vert c\Vert_2$ for vectors or the Frobenius norm $\Vert\cdot\Vert$, introduced in Section \ref{sec:notation}, for matrices and tensors. As shown in \cite{derksen2018general}, the denoising of a vector $c$ via a decomposition $c= a + b$ that simultaneously minimizes the values $\Vert a \Vert_X$ and $\Vert b \Vert_Y$ is governed by the \textit{Pareto frontier}, which models the competing objectives of minimizing the two norms in terms of Pareto efficiency, and the above $XY$\textit{-decomposition} that achieves this is deemed \textit{Pareto efficient}. Moreover, \cite{derksen2018general} defines the related notion of the \textit{Pareto subfrontier}, which relates the three norms $\Vert\cdot \Vert_X$, $\Vert \cdot\Vert_Y$, $\Vert\cdot\Vert_2$ and their induced decompositions $XY$, $X2$, and $2Y$, describing the conditions under which these decompositions can achieve Pareto efficiency.

\subsection{Amplification-based Denoising}
To make use of the denoising framework introduced in \cite{derksen2018general} requires the calculations of various norms for the vectors of interest. While the Frobenius norm of a tensor is easily obtained, computing either the nuclear norm \cite{friedland2018nuclear} or the spectral norm \cite{hillar2013most} for tensors is NP-hard. In order to obtain an approximation to the tensor spectral norm, the co-authors developed the methodology of \textit{tensor amplification} \cite{tokcan2021algebraic}. For a matrix $A$ with singular values $\lambda_1,\ldots,\lambda_r$, the function $\phi: A\rightarrow AA^{\intercal}A$ produces the matrix $AA^{\intercal}A$ whose singular values are $\lambda_1^3,\ldots,\lambda_r^3$. Repeated applications of $\phi(\cdot)$ will \textit{amplify} the larger singular values, which correspond to the sparse or low rank components of the matrix, while minimizing smaller singular values that likely correspond to noise.

Analogously, tensor amplification utilizes degree $d$ polynomial functions on tensors to amplify the low rank structure. Moreover, for each amplification map $\Phi_{\sigma'}$ there exists a corresponding norm $\Vert\cdot\Vert_{\sigma',d}$ that approximates the tensor spectral norm, in the sense that $\lim_{d\rightarrow\infty}\Vert\mathcal{T}\Vert_{\sigma',d}= \Vert\mathcal{T}\Vert_{\sigma}$. Two such amplification maps -- $\Phi_{\sigma,4}$ and $\Phi_{\#}$ -- were introduced for order 3 tensors in \cite{tokcan2021algebraic}, with $\Phi_{\#}$ being show to be a better approximation to the tensor spectral norm than $\Phi_{\sigma,4}$.

Using these amplification maps we developed the method of \emph{amplification-based tensor denoising}, presented in Algorithm \ref{alg:amp} below. Unlike in typical denoising methods, the algorithm can effectively remove noise from the input tensor without resorting to computations on its flattenings. The algorithm utilizes the $2Y$-decomposition framework of \cite{derksen2018general} and the tensor spectral norm approximations $\Phi$ to denoise a given tensor $\mathcal{T}$. The algorithm allows for the choice of amplification map as well as the number of amplifications per round. For third order tensors the amplification map  $\Phi_{\#}$ was used, while for fourth order tensors a compatible version of $\Phi_{\sigma,4}$ was employed as there is no currently known analogue of the $\Phi_{\#}$ map for fourth order tensors. Multiple experiments were performed with $m$, the number of amplifications per round, ranging from 1 to 10, with $m=5$ being found to produce the best denoising performance. 

\begin{algorithm}
\begin{algorithmic}
\State{$\mathcal{D}\gets$ \Call{Denoise\_Amplification}{$\mathcal{T},\Phi,m$}}
	\State {$\epsilon \gets \Vert\mathcal{T}\Vert$}
	\State {$\mathcal{N} \gets \mathcal{T}$}
	\While {true}
		\State {$\mathcal{A} \gets \Phi^m(\mathcal{N})$}
		\State {$\mathcal{A} \gets \frac{ \mathcal{A} }{\Vert \mathcal{A} \Vert}$}
		\State {$\mathcal{N} \gets \mathcal{N}-\langle \mathcal{A},\mathcal{N}\rangle \mathcal{A}$}
		\State {Break if $\Vert \mathcal{N}\Vert < \epsilon$}
	\EndWhile
	\State {$\mathcal{D} \gets \mathcal{T}-\mathcal{N}$}
\end{algorithmic}
\caption{Amplification-based tensor denoising.}
\label{alg:amp}
\end{algorithm}

\subsection{Stable Slice Rank Denoising}
As noted in Section \ref{sec:decompdenoising}, a degenerate tensor $\mathcal{T}$ of rank $r$ may not have a best rank $k<r$ approximation for a given rank $k$. In such cases, a tensor may be approximated to any desired precision by rank $j<k$ tensors. This is problematic for iterative approaches to tensor approximation, such as in the decomposition-based approaches to tensor denoising. This is due to the set of all tensors for a given rank $r$ not being closed with respect to the Zariski topology \cite{de2008tensor}. Slice rank is an alternative definition of tensor rank that was introduced in \cite{blasiak2016cap} in relation to the cap set problem. Unlike the regular notion of tensor rank, slice rank is Zariski closed, thus can circumvent the issue of tensor degeneracy. Following Tao \cite{slicerank}, the \textit{slice rank} of a tensor $\mathcal{T}$ is the least non-negative integer $srk$ such that $\mathcal{T}$ is a sum of tensors  with slice rank 1, i.e., $\mathcal{T}=\sum_{i=1}^r \mathcal{T}_i$, where $\mathcal{T}_i$ is contained in the tensor product space
\begin{equation}
    V_1\circ\cdots V_{i-1}\circ s\circ V_{i-1}\circ\cdots V_d, 
\end{equation}
where $V_j$ are vector spaces and $s$ is a vector in some $V_i$.

In \cite{rudelson2007sampling}, the notion of a \textit{stable rank} for matrices was introduced, in which the matrix rank function  $rank(A)$, is replaced by the \emph{numerical rank function}, $\frac{\Vert A\Vert^2}{\Vert A\Vert_{\sigma}^2}$, or the related \emph{stable nuclear rank} $\frac{\Vert A\Vert_{\star}^2}{\Vert A\Vert^2}$. These ranks are stable in the sense that small perturbations of the values of the matrix $A$ will not change their value. Extending this methodology to tensors, we can define the \textit{stable slice rank} as
\begin{equation}
    ssrk(\mathcal{T})=\dfrac{(\sum_{i=1}^d \Vert T_{(i)}\Vert_{\star})^2}{\Vert \mathcal{T}\Vert^2},
\end{equation}
where $\mathcal{T}_{(i)}$ are the mode-$i$ flattenings of $\mathcal{T}$. By estimating the stable slice rank of a noisy tensor, we can devise a new decomposition-based denoising method -- \textit{stable slice rank (SliceRank)} denoising, presented in Algorithm \ref{alg:slicerank} below.

In this method, a tensor is denoised by finding a decomposition $\mathcal{T}=\mathcal{D}+\mathcal{N}$ that minimizes the Frobenius norm of $D=\sum_{i=1}^d \mathcal{S}_i$ under the constraints that the nuclear norms of the flattenings of $\mathcal{N}$ are all $\leq \lambda$, a user-defined hyperparameter. The method also simultaneously minimizes the sum of the nuclear norms of $S_{(i)}$, the mode-$i$ flattening of $\mathcal{S}_i$. Typically, the $S_{(i)}$ produced by the algorithm will have low (matrix) rank. 
\begin{algorithm}
\begin{algorithmic}

\State{($\mathcal{D},\{\mathcal{S}_i\},ssrk)\gets$ \Call{Denoise\_SliceRank}{$\mathcal{T},\lambda,acc$}}
    \State {$\mathcal{S}_i \gets \mathbf{0}\in \R^{p_1\times p_2 \times \cdots \times p_d}$ } 
	\State {$curr\_acc \gets 0$}
	\While {$curr\_acc < acc$}
        \For {$i\gets 1:d$}
    		\State {$\mathcal{A} \gets \mathcal{T}-\sum_{j\neq i} S_j$}
            \State {$\mathbf{q} \gets$ \Call{circshift}{$[1,\ldots,d],-(i-1)$}}
    		\State {$\mathcal{A} \gets \mathcal{A}^{\langle \mathbf{q} \rangle}$}
            \State {$(U,D,V) \gets$ \Call{svd}{$\mathcal{A}_{(i)}$} }
    		\State {$E_i \gets$ \Call{max}{$D-\lambda,0$} }
    		\State {$F \gets U\cdot E_i\cdot V^{T}$}
    		\State {$\mathcal{F} \gets$ \Call{reshape}{$F,p_i,\ldots,p_d,p_1,\ldots,p_{i-1}$} }
            \State {$\mathbf{q} \gets$ \Call{circshift}{$[1,\ldots,d],(i-1)$}}
    		\State {$\mathcal{S}_i \gets \mathcal{F}^{\langle \mathbf{q} \rangle}$}
        \EndFor
        \State{$\mathcal{D}_{curr} \gets \sum_{i=1}^d \mathcal{S}_i$}
        \State{$\mathcal{N} = \mathcal{T}-\mathcal{D}_{curr}$}
        \State {$curr\_acc \gets \dfrac{\langle\mathcal{N},\mathcal{D}_{curr}\rangle}
                {\lambda\sum_{j=1}^d \Vert S_{(i)}\Vert_{\star} }$}
    \EndWhile
	\State {$\mathcal{D} \gets \sum_{i=1}^d\mathcal{S}_i$}
    \State {$ssrk \gets \dfrac{(\sum_{i=1}^d \Vert S_{(i)}\Vert_{\star})^2}{\Vert \mathcal{D}\Vert^2}$ }
\end{algorithmic}
\caption{Stable SliceRank denoising.}
\label{alg:slicerank}
\end{algorithm}

Algorithm \ref{alg:slicerank} depicts the implementation of SliceRank denoising, which utilizes a number of auxiliary functions from MATLAB \cite{MATLAB2022a}: \texttt{circshift} performs a cyclic permutation of an index set $[1,\ldots,d]$, with the second parameter determining the number of forward or backwards shifts; \texttt{reshape} is used to flatten a tensor into a matrix with the specified dimensions; and \texttt{diag} returns a vector comprising the entries on the main diagonal of the specified matrix. The algorithm takes as hyperparameters $\lambda$ as described above and $acc\in(0,1]$, the specified accuracy level that once achieved the algorithm terminates. The algorithm returns the denoised tensor $\mathcal{D}$, the decomposition factors $\mathcal{S}_i$, and $ssrk$, the stable slice rank of $\mathcal{D}$. The hyperparameters were optimized via grid search over the ranges $\lambda\in\{10^{-2},0.1,1,10\}$ and $acc\in\{0.90,0.95\}$.

\FloatBarrier

\subsection{Stable $X$-Rank Denoising}
 In Derksen \cite{derksen2022g}, the $G$-stable rank of a tensor was introduced that, like slice rank, is Zariski closed. Thus, every tensor $\mathcal{T}$ has a best $G$-stable rank $r$ approximation. While $G$-stable rank is directly proportional to slice rank, it has a number of computational advantages, namely that it is additive (rather than sub-additive in the case of slice rank) and super-multiplicative \cite{derksen2022g}. Moreover, $G$-stable rank can take on fractional values, potentially providing greater discriminating power in using it to evaluate approximations to given tensor. The \textit{$G$-stable $\alpha$ rank} of a tensor can be defined as
\begin{equation}
rk_{\alpha}^G(\mathcal{T})=\sup_{g\in G}\min_i\dfrac{\alpha_i\Vert g\cdot\mathcal{T}\Vert^2}{\Vert\left(g\cdot\mathcal{T}\right)_{(i)}\Vert_{\sigma}^2},
\end{equation}
where $\alpha=(\alpha_1,\ldots,\alpha_d)$ and $g$ is an element of a reductive group $G$, i.e., $g\in SL(\R^{p_1})\times \cdots \times SL(\R^{p_d})$.

Using the above definition we can define the related concept of \textit{stable $X$-rank}, which is
\begin{equation}
   sxrk^G(\mathcal{T})=\max_{\alpha} rk_{\alpha}^G(\mathcal{T}),
\end{equation}
where $\alpha$ is subject to the restriction that $\sum_i\alpha_i = d$. Algorithm \ref{alg:xrank} depicts the implementation of the \textit{stable} $X$\textit{-Rank (XRank)} denoising method. Like SliceRank, the method imposes a constraint on the nuclear norm of the flattenings of $\mathcal{N}$. However, in the XRank method, this cutoff is determined automatically using Algorithm \ref{alg:cutoff}. The hyperparameters were optimized via grid search over the ranges $\lambda\in\{10^{-2},0.1,1,10\}$ and $acc\in\{0.90,0.95\}$.
\begin{algorithm}
\begin{algorithmic}
\State{$c\gets$ \Call{Find\_Cutoff}{$\mathbf{f}=[\lambda_1,\ldots,\lambda_r]^{\intercal},\lambda$}}
    \State{$r \gets |f|$}
    \State{$\mathbf{t} \gets \mathbf{0}\in\R^r$}
    \For {$i\gets 1:r$}
        \State {$t_i \gets \lambda\dfrac{\sum_{j=1}^i\lambda_j}{1+\lambda\cdot i}$}
    \EndFor
    \State{$S \gets \mathbf{0}\in\R^{r\times r}$}
    \For {$i\gets 1:r$}
        \For {$j\gets 1:r$}
            \State{$s_{ij} \gets$ \Call{max}{$f_i-t_j,0$}}
        \EndFor
    \EndFor
    \State{$v \gets \mathbf{0}\in\R^r$}
    \For {$j\gets 1:r$}
        \State{$v_j\gets \sum_{i=1}^r (f_{ij}-s_{ij})^2+\lambda\sum_{i=1}^r (s_{ij})^2$}
    \EndFor
    \State{$k \gets$ \Call{argmin}{$v$}}
    \State{$c \gets t_k$}
\end{algorithmic}
\caption{Determine the nuclear norm cutoff for XRank denoising.}
\label{alg:cutoff}
\end{algorithm}

\begin{algorithm}
\begin{algorithmic}
\State{($\mathcal{D},\{\mathcal{S}_i\},sxrk)\gets$ \Call{Denoise\_XRank}{$\mathcal{T},\lambda,acc$}}
    \State {$\mathcal{S}_i \gets \mathbf{0}\in \R^{p_1\times p_2 \times \cdots \times p_d}$ } 
	\State {$curr\_acc \gets 0$}
	\While {$curr\_acc < acc$}
        \For {$i \gets 1:d$}
    		\State {$\mathcal{A} \gets \mathcal{T}-\sum_{j\neq i} S_j$}
            \State {$\mathbf{q} \gets$ \Call{circshift}{$[1,\ldots,d],-(i-1)$}}
    		\State {$\mathcal{A} \gets \mathcal{A}^{\langle \mathbf{q} \rangle}$}
    		\State {$(U,D,V) \gets$ \Call{svd}{$\mathcal{A}_{(i)}$} }
            \State {$c \gets$ \Call{Find\_Cutoff}{\Call{diag}{$D$},$\lambda$}}
    		\State {$E_i \gets$ \Call{max}{$D-c,0$} }
    		\State {$F \gets U\cdot E_i\cdot V^{T}$}
    		\State {$\mathcal{F} \gets$ \Call{reshape}{$F,p_i,\ldots,p_d,p_1,\ldots,p_{i-1}$} }
            \State {$\mathbf{q} \gets$ \Call{circshift}{$[1,\ldots,d],(i-1)$}}
    		\State {$\mathcal{S}_i \gets \mathcal{F}^{\langle \mathbf{q} \rangle}$}
        \EndFor
        \State{$\mathcal{D}_{curr} \gets \sum_{i=1}^d \mathcal{S}_i$}
        \State{$\mathcal{N} = \mathcal{T}-\mathcal{D}_{curr}$}
        \State{$y \gets 0$}
        \For {$i \gets 1:d$}
            \State {$\mathbf{q} \gets$ \Call{circshift}{$[1,\ldots,d],-(i-1)$}}
    		\State {$\mathcal{B} \gets \mathcal{N}^{\langle \mathbf{q} \rangle}$}
    		\State {$(U,D,V) \gets$ \Call{svd}{$\mathcal{B}_{(i)}$} }
            \State {$y\gets y + d_1^2$} \Comment{$d_1$ is the largest singular value of $D$}
        \EndFor
        \State {$y\gets \sqrt{y}$}
        \State {$curr\_acc \gets \dfrac{\langle\mathcal{N},\mathcal{D}_{curr}\rangle}
                {y\cdot\sqrt{ \sum_{j=1}^d \Vert S_{(j)}\Vert_{\star}^2} }$}
    \EndWhile
	\State {$\mathcal{D} \gets \sum_{i=1}^d\mathcal{S}_i$}
    \State {$sxrk \gets \dfrac{\sum_{j=1}^d \Vert S_{(j)}\Vert_{\star}^2}{\Vert\mathcal{D}\Vert^2}$ }
\end{algorithmic}
\caption{Stable XRank denoising.}
\label{alg:xrank}
\end{algorithm}
\section{Experimental Results and Discussion}
In order to evaluate the various denoising methods under consideration, two sets of synthetic tensors were generated with varying orders, ranks, and dimensions, resulting in 512 parameter combinations. For each combination, one hundred (100) tensors were generated.  For all synthetic tensors, varying amounts of noise were added from a standard Gaussian distribution $\mathcal{N}(0,1)$, with the resulting noisy tensors having signal-to-noise ratios (SNR) ranging from 20 dB to $-20$ dB. The full range of parameters is provided in Table \ref{tab:tensorparams}.

\begin{table}[h]
    \centering
    \caption{Parameters and their respective values used to generate the synthetic tensor datasets.}
    \begin{tabular}{|c|c|}
    \hline
    \textbf{Parameter} & \textbf{Range/Values}\\
    \hline
    Distribution & Normal $\mathcal{N}(0,1)$\\
    Order & $3,4$\\
    Rank & $[1,5], 10,20,25$\\
    Size & $5,10,25,50$\\
    SNR & $20,10,5,1,-1,-5,-10,-20$\\
    \hline
    \end{tabular}
    \label{tab:tensorparams}
\end{table}
Additionally, two sets of tensors were extracted from electrocardiogram (ECG) signals to which Gaussian noise was added prior to tensor extraction, using the same range of resultant SNRs as those employed in the generation of the synthetic tensors.

\subsubsection{Uniform Synthetic Tensors} In this dataset, the dimensions of a given tensor are chosen uniformly across each mode. To generate synthetic tensors from a distribution $\mathcal{D}$ of a given rank $r$, size $s$, and order $d$, scalar values $\lambda_1,\ldots,\lambda_r$ are chosen from $\mathcal{D}$, then for each mode $j$, $r$ random vectors $x_{j,i}\in\mathbb{R}^s$ are chosen from $\mathcal{D}$. The synthetic tensor is then
$$
\sum_{i=1}^r \lambda_i x_{1,i} \circ x_{2,i} \circ \cdots \circ x_{d,i}.
$$

\subsubsection{Non-Uniform Synthetic Tensors} In this dataset, one mode $m_k$ of a given tensor is ``stretched" to a different dimension $d_k$ by choosing a number uniformly in the range $d_k=[s,min(500,s^d)]$, i.e., the lower bound is the dimension of the other models while the upper bound is the product of the dimensions of each mode or 500, whichever is lower.  After choosing the stretch mode and its dimension the tensors are generated in the same manner as for the uniform tensors above.

\subsubsection{ECG Waveform Tensors} The PTB Diagnostic ECG Database \cite{bousseljot1995nutzung} is comprised of high resolution (1 kHz) digitized recordings of electrocardiograms (ECGs) from patients with various cardiovascular diseases, including myocardial infarction, heart failure, and arrhythmia, as well as healthy controls. The database is publicly available via PhysioNet \cite{PhysioNet}.

Tensor-based methods have been shown to be effective for a number of ECG analytical tasks, a survey of such methods can be found in \cite{padhy2020power}. Given the utility of tensor-based methods in this context and that such that recordings of physiological signals may be corrupted by noise yields a natural application of the proposed denoising methods. In order to evaluate these methods, we first must construct tensors from the ECGs. In forming these tensors, one has to consider the amount of signal over which to perform subsequent signal processing and feature extraction: two methods were employed. In the first, ninety (90) seconds of a patient's ECG recording was sampled across all twelve ECG leads, while in the second method three windowed samples of thirty (30) seconds each were extracted.

Using these two sampling strategies we adapted the tensor formation method introduced in \cite{hernandez2021multimodal} that has been shown to be effective for subsequent applications of machine learning for prognosticating severe cardiovascular conditions \cite{kim2022prediction,mathis2022prediction}. In this method, each ECG signal is preprocessed using the taut string method, which produces a piecewise linear approximation of a given signal, parametrized by $\epsilon$, which controls the coarseness of the approximation. Given a discrete signal $x=(x_1,\ldots,x_n)$ one can define the finite difference $D(x)=(x_2-x_1,\ldots,x_n-x_{n-1})$. For a fixed $\epsilon>0$, the taut string estimate of $x$ is the unique function $y$ such that $\Vert x-y\Vert_{\infty}\leq\epsilon$ and $\Vert D(y)\Vert_2$ is minimal. The taut string approximation can be found efficiently using the method in \cite{davies2001local}.

After the taut string approximation for a given signal is found, six morphological and statistical features are extracted following \cite{hernandez2021multimodal}. This process is repeated for five values of epsilon: $(0.0100, 0.1575, 0.3050, 0.4525, 0.6000)$. As each patient's ECG recording is comprised of the standard 12 leads, the approximation of each 90 second ECG sample via taut string and the extraction of taut string features yields third order tensors of size $5\times6\times12$ for each patient. For the windowed samples, fourth order tensors were formed of size $5\times 6\times 12\times 3$, with the fourth mode corresponding to the features extracted in each window.

\subsubsection{Adding Noise} For every generated synthetic tensor, a set of noisy tensors was created by adding Gaussian noise ($\mathcal{N}(0,1)$) so that the resultant tensors had SNRs in the range $[20,10,5,1,-1,-5,-10,-20]$. For the ECG waveform tensors, Gaussian noise was added to each ECG signal to produce a set of noisy signals with the same SNR range as for the synthetic tensors. However, this was performed prior to tensor formation given that in practical applications ECG signals themselves may come with some intrinsic amount of noise, rather than noise being introduced to the tensors directly.

\begin{table*}[t!]
\centering
\caption{Mean (SD) SNR, in decibels, after tensor denoising across all parameters.}
\begin{tabular}{|c|cccccc|}
\hline
\textbf{Starting SNR} &          \textbf{HOOI} &            \textbf{ALS} &         \textbf{Wiener} &            \textbf{Amp} &     \textbf{SliceRank} &         \textbf{XRank} \\
\hline
 20 &  19.57 (3.32) &   \textit{28.67 (11.1)} &  \textbf{29.05 (13.19)} &   10.0 (14.13) &  18.79 (1.59) &  10.89 (5.64) \\
 10 &  10.59 (1.12) &   \textit{19.91 (8.88)} &   \textbf{20.22 (10.1)} &   8.65 (11.09) &  13.21 (2.21) &   9.84 (4.15) \\
 5  &   5.81 (0.75) &   \textbf{15.11 (8.26)} &   \textit{14.94 (8.59)} &    7.85 (9.64) &   8.18 (2.52) &   8.56 (3.13) \\
 1  &    1.87 (0.7) &   \textbf{11.12 (7.98)} &   \textit{10.34 (7.64)} &    7.01 (8.56) &   3.54 (2.25) &   6.91 (2.49) \\
-1  &   -0.12 (0.7) &     \textbf{9.1 (7.88)} &    \textit{7.96 (7.01)} &     6.48 (8.1) &   1.18 (2.03) &   5.86 (2.36) \\
-5  &  -4.12 (0.69) &    \textit{4.99 (7.77)} &    3.37 (5.69) &    \textbf{5.17 (7.45)} &  -3.46 (1.57) &    3.24 (2.6) \\
-10 &  -9.12 (0.69) &   \textit{-0.19 (7.65)} &   -2.25 (4.57) &    \textbf{2.85 (7.35)} &  -9.02 (1.12) &  -0.58 (3.43) \\
-20 &  -19.11 (0.7) &  \textit{-10.38 (7.39)} &  -11.17 (7.38) &  -11.79 (6.98) &  -19.61 (0.6) &  \textbf{-9.22 (4.79)} \\
\hline
\multicolumn{7}{c}{(a) Uniformly sized tensors.}\\
\hline
\textbf{Starting SNR} &          \textbf{HOOI} &            \textbf{ALS} &         \textbf{Wiener} &            \textbf{Amp} &     \textbf{SliceRank} &         \textbf{XRank} \\
\hline
 20 &   23.91 (7.33) &   \textbf{30.81 (12.7)} &   \textit{29.46 (13.3)} &  11.68 (14.81) &    18.7 (1.53) &  11.85 (5.83) \\
 10 &   15.45 (4.71) &  \textbf{22.75 (10.29)} &  \textit{21.39 (10.79)} &  10.33 (11.77) &   13.28 (2.11) &  10.75 (4.34) \\
 5  &   10.99 (3.94) &   \textbf{18.25 (9.55)} &   \textit{16.28 (9.86)} &   9.45 (10.34) &    8.26 (2.44) &     9.4 (3.3) \\
 1  &    7.27 (3.67) &   \textbf{14.49 (9.23)} &   \textit{11.84 (9.03)} &    8.25 (9.36) &    3.63 (2.19) &    7.68 (2.6) \\
-1  &    5.38 (3.63) &   \textbf{12.57 (9.12)} &    \textit{9.54 (8.41)} &    7.32 (9.03) &    1.28 (2.01) &    6.6 (2.42) \\
-5  &     1.5 (3.64) &    \textbf{8.66 (8.99)} &    \textit{5.38 (6.92)} &    4.89 (9.02) &   -3.37 (1.61) &   3.95 (2.56) \\
-10 &   -3.51 (3.61) &    \textbf{3.65 (8.86)} &    0.08 (5.37) &    \textit{1.65 (9.72)} &   -8.92 (1.29) &  -0.13 (3.39) \\
-20 &  -13.59 (3.51) &    \textbf{-6.6 (8.57)} &   \textit{-7.94 (7.44)} &  -11.74 (8.56) &  -19.56 (0.64) &  -9.01 (4.66) \\
\hline
\multicolumn{7}{c}{(b) Non-uniformly sized tensors.}
\end{tabular}
\label{tab:overallsnr}
\end{table*}
\begin{figure*}[hbt]
    \begin{subfigure}{0.5\textwidth}
    \includegraphics[width=\textwidth]{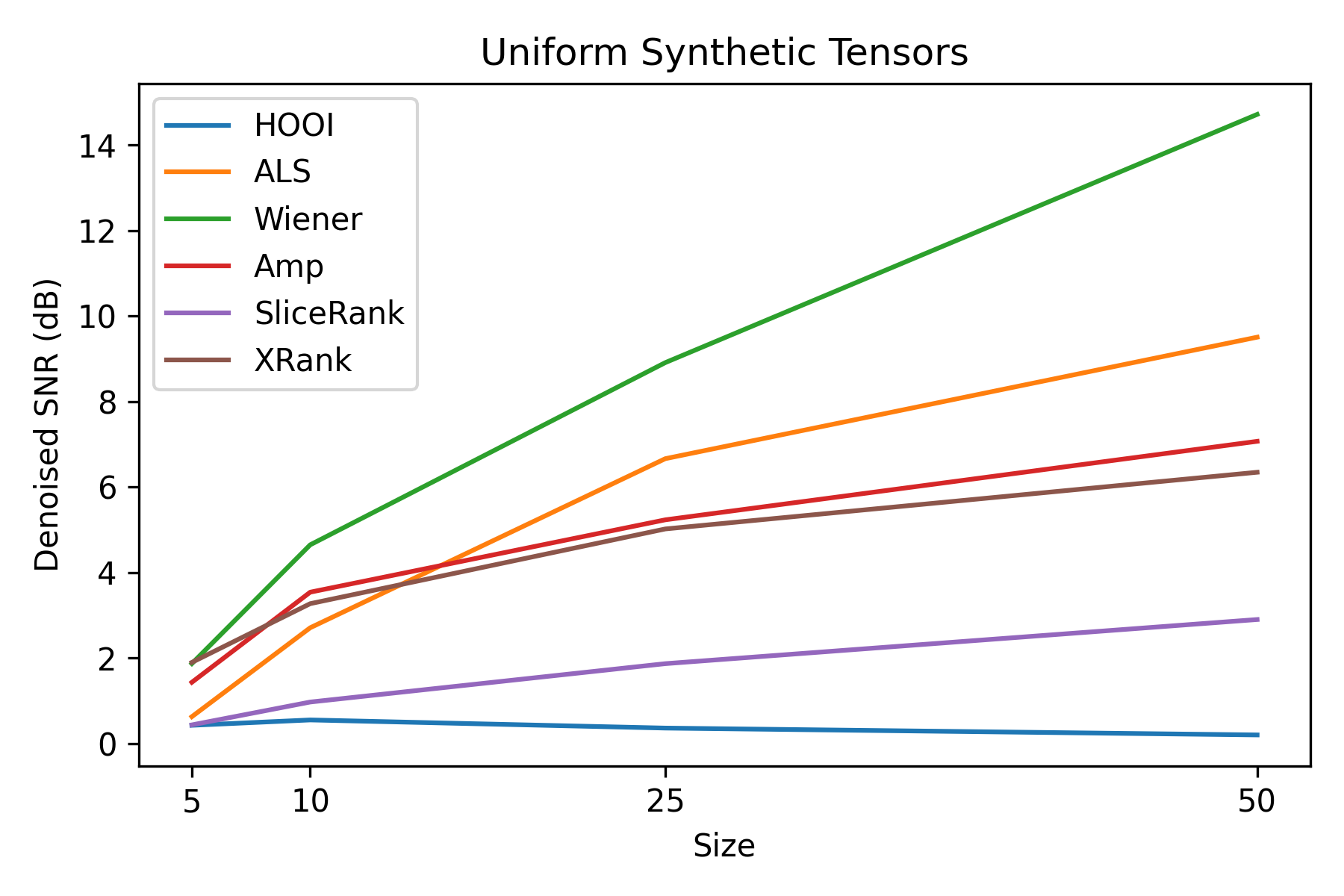}
    \caption{}
    \end{subfigure}
    \begin{subfigure}{0.5\textwidth}
    \includegraphics[width=\textwidth]{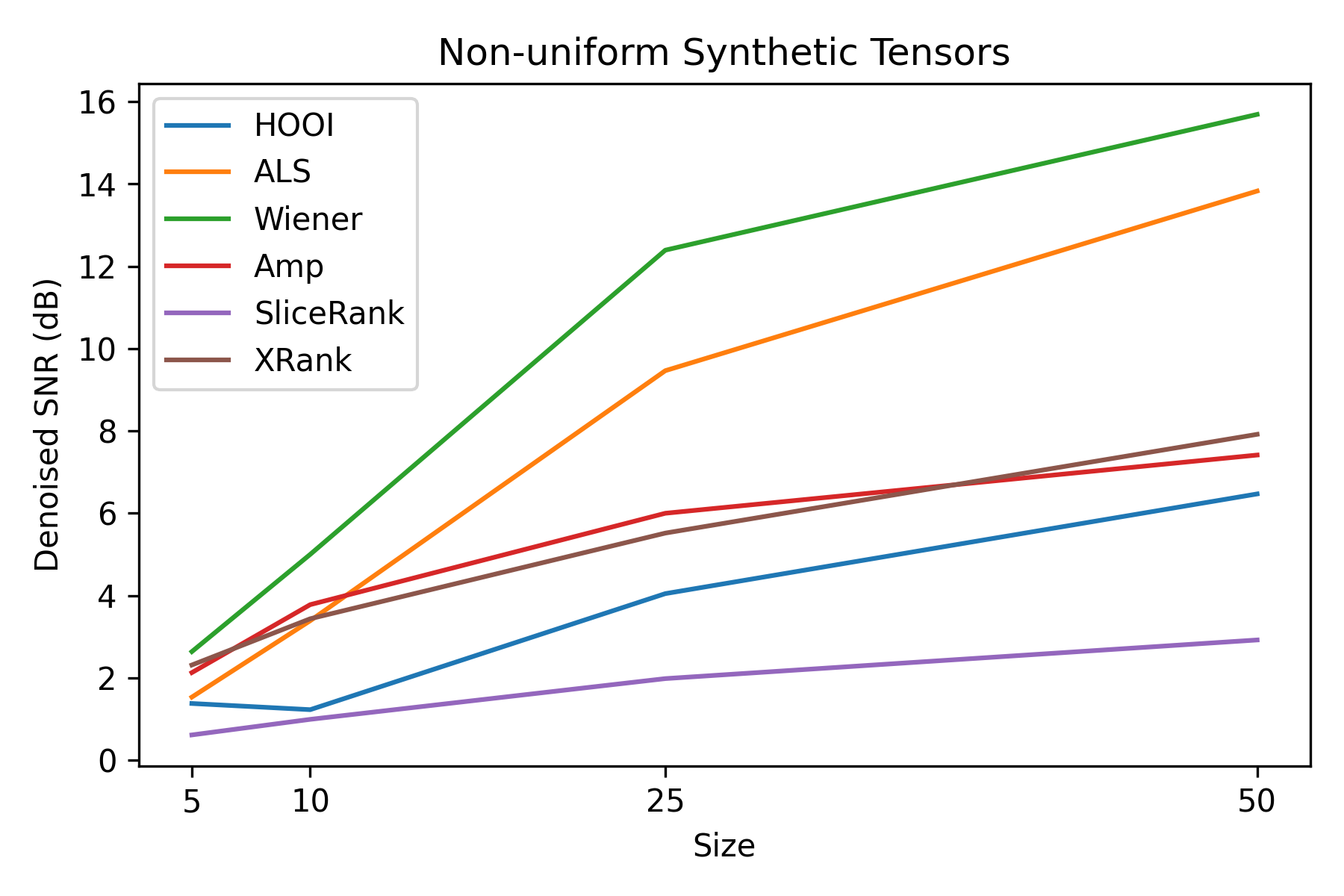}
    \caption{}
    \end{subfigure}
    \caption{Denoising performance with respect to tensor size.}
    \label{fig:synthetic-sizes}
\end{figure*}
\subsection{Results: Synthetic Data}
The overall denoising performance for third order tensors across various ranks and tensors sizes are presented in Table \ref{tab:overallsnr}. The performance statistics for only one order are presented due to the incomparable sizes of the non-uniform tensors across orders, please see Table \ref{tab:4overallsnr} in Appendix \ref{sec:order4results} for the fourth order 4. The best performing denoising algorithms for uniformly sized tensors, as depicted in Table \ref{tab:overallsnr} (a), varied by noise level. For cleaner tensors (20 and 10 dB), the multiway Wiener filter performed best overall, achieving mean and standard deviations in denoised SNRs of 20.95 (13.19) and 20.22 (10.1) dBs, respectively. For moderately noisy tensors (5, 1, and $-1$ dB), ALS was the best performing denoising method, achieving denoised SNRs of 15.11 (8.26), 11.12 (7.98), and 9.1 (7.88) dBs. For nosier tensors with starting SNRs of $-5$ and $-10$, tensor amplification produced the best denoised SNRs of 3.37 (5.69) and $-2.25$ (4.57) dBs. Finally for tensors with starting SNRs of $-20$ dB, the noisiest tensors evaluated, XRank produced on average the highest denoised SNR of $-9.22$ (4.79) dB. The results for non-uniformly sized tensors, as depicted in Table \ref{tab:overallsnr} (b), are much clearer, with ALS achieving the best denoised SNRs across all starting SNRs, ranging from 30.81 (12.7) dBs for tenors with starting SNRs of 20 dB to $-6.6$ (8.57) dB for the noisiest tensors (starting SNRs of $-20$ dB).

The relationship between tensor \textit{size} (dimension of each mode) and achieved denoised SNRs is depicted in Figure \ref{fig:synthetic-sizes}. With the exception of HOOI, all other denoising algorithms see improvements in achieved SNR as the size of the tensor increases. The multiway Wiener filter maintains the best denoising performance as size increases, followed by ALS. Both amplification and XRank have similar denoising performances, while slice rank and HOOI having the lowest performance overall.
\begin{figure*}[hbt]
    \begin{subfigure}{0.5\textwidth}
    \includegraphics[width=\textwidth]{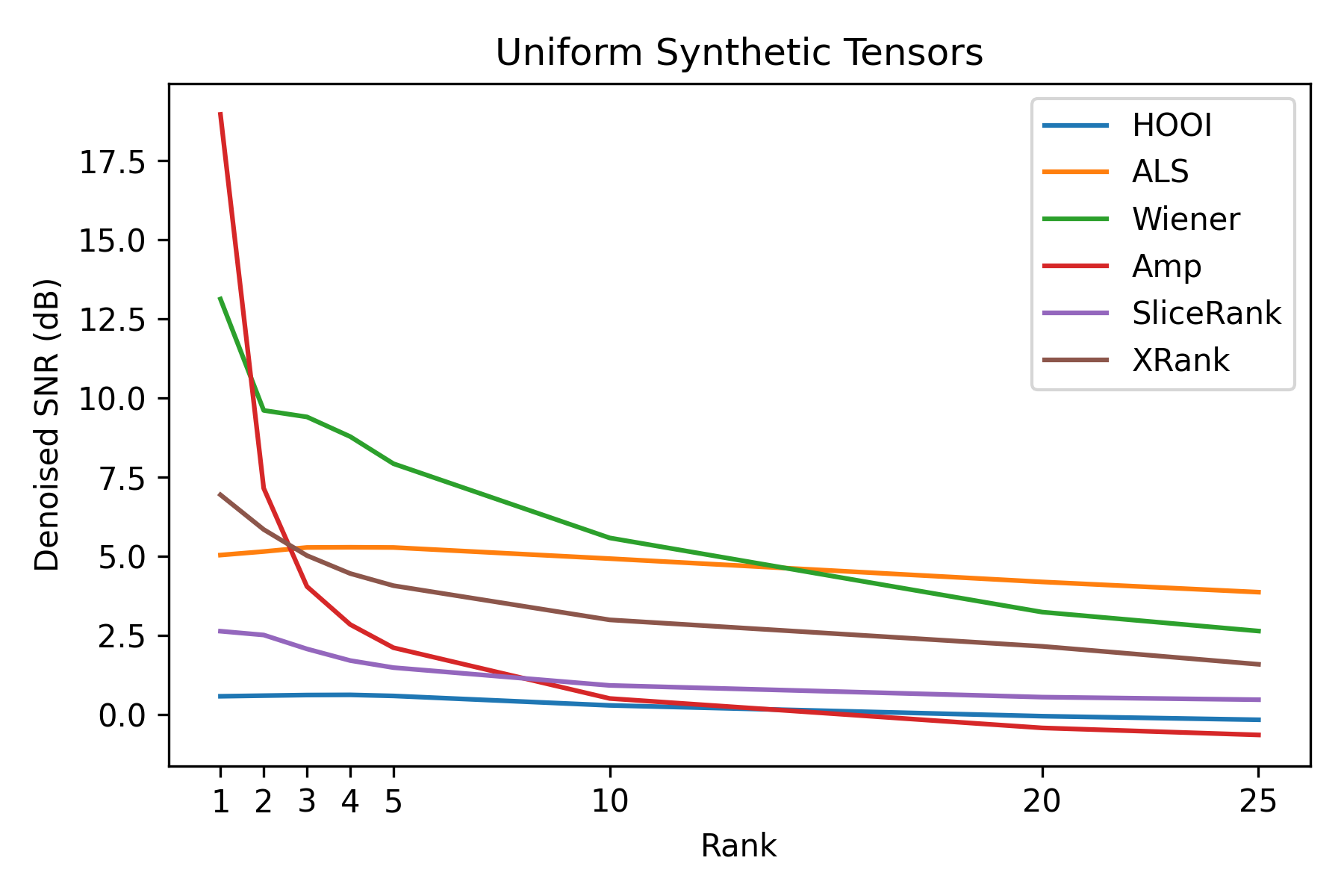}
    \caption{}
    \end{subfigure}
    \begin{subfigure}{0.5\textwidth}
    \includegraphics[width=\textwidth]{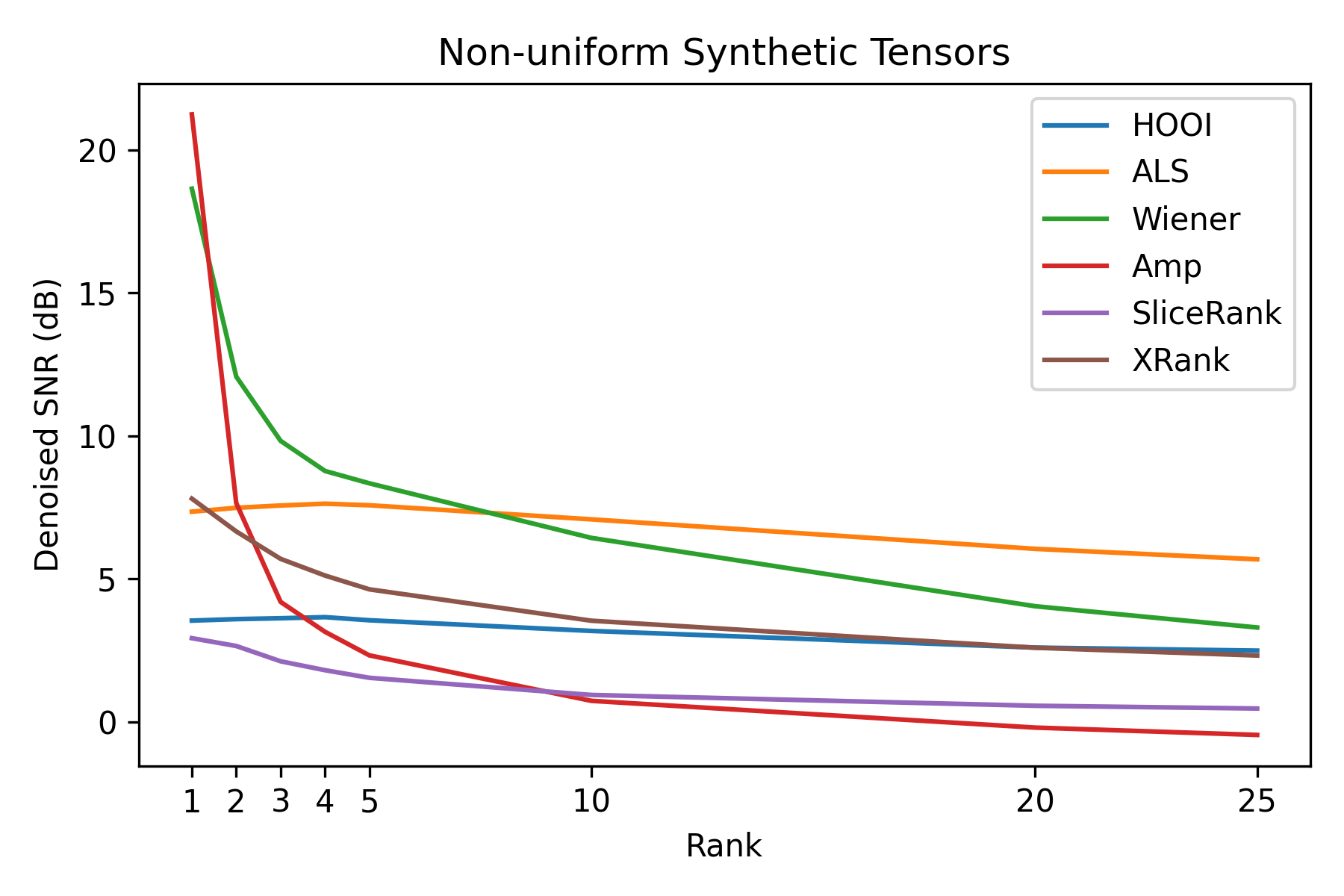}
    \caption{}
    \end{subfigure}
     \caption{Denoising performance with respect to tensor rank.}
   \label{fig:synthetic-ranks}
\end{figure*}

\begin{table*}[hbt]
\centering
\caption{Mean (SD) denoised SNR, in decibels, for low rank and noisy tensors.}
\begin{tabular}{|c|cccccc|}
\hline
\textbf{Starting SNR} &          \textbf{HOOI} &            \textbf{ALS} &         \textbf{Wiener} &            \textbf{Amp} &     \textbf{SliceRank} &         \textbf{XRank} \\
\hline
1 & 1.59 (0.36) & 5.97 (2.93) & \textit{12.21 (4.05)} & \textbf{14.98 (7.72)} & 5.64 (2.29) & 8.89 (1.91) \\
-1 & -0.42 (0.35) & 3.92 (2.92) & \textit{10.31 (3.85)} & \textbf{13.66 (7.28)} & 3.33 (2.11) & 7.34 (2.01) \\
-5 & -4.43 (0.32) & -0.17 (2.91) & \textit{6.42 (3.78)} & \textbf{10.68 (6.90)} & -1.52 (1.70) & 3.77 (2.34) \\
-10 & -9.45 (0.30) & -5.24 (2.93) & \textit{1.90 (3.84)} & \textbf{6.16 (7.47)} & -7.49 (1.39) & -0.79 (2.88) \\
-20 & -19.45 (0.30) & -15.29 (2.91) & \textit{-8.86 (5.90)} & \textbf{-4.36 (7.84)} & -18.79 (1.03) & -10.31 (3.48) \\
\hline
\multicolumn{7}{c}{(a) Uniformly sized tensors.}\\
\hline
\textbf{Starting SNR} &          \textbf{HOOI} &            \textbf{ALS} &         \textbf{Wiener} &            \textbf{Amp} &     \textbf{SliceRank} &         \textbf{XRank} \\
\hline
1 & 4.59 (1.82) & 8.32 (4.32) & \textbf{16.96 (8.56)} & \textit{16.15 (8.84)} & 5.85 (2.30) & 9.68 (1.96) \\
-1 & 2.58 (1.82) & 6.25 (4.30) & \textbf{14.99 (8.41)} & \textit{14.88 (8.34)} & 3.51 (2.24) & 8.25 (2.03) \\
-5 & -1.46 (1.84) & 2.12 (4.29) & \textit{10.77 (8.21) }& \textbf{12.06 (7.72)} & -1.25 (2.06) & 4.89 (2.39) \\
-10 & -6.50 (1.87) & -2.98 (4.31) & \textit{4.07 (6.49)} & \textbf{7.84 (7.93)} & -7.19 (1.87) & 0.21 (3.09) \\
-20 & -16.52 (1.87) & -13.05 (4.31) & \textit{-7.18 (6.61)} & \textbf{-1.79 (9.00)} & -18.86 (0.84) & -9.05 (3.87) \\
\hline
\multicolumn{7}{c}{(b) Non-uniformly sized tensors.}
\end{tabular}
\label{tab:lowranklownsr}
\vspace{-2em}
\end{table*}

The relationship between tensor \textit{rank} and achieved denoised SNRs is depicted in Figure \ref{fig:synthetic-ranks}. Tensor amplification achieves the best rank-1 performance for both uniformly and non-uniformly sized tensors, followed by the multiway Wiener filter. The denoising performance of both methods decreases as tensor rank increases, with the multiway Wiener filter maintaining denoising performance for higher ranks than amplification. ALS has lower performance at low ranks but generally maintains its denoising performance as rank increases, ultimately achieving the best results by rank 20. XRank has greater performance than SliceRank for uniformly sized tensors, but their denoising performances converge prior to rank 20 for non-uniformly sized ones. HOOI has the lowest denoising performance for uniformly sized tensors, but performs slightly better than the amplification and SliceRank methods for non-uniformly sized tensors with ranks greater than 5.

The results depicted in Table \ref{tab:lowranklownsr} provide a further investigation into the denoising performance for low rank (ranks 1 and 2) and high noise (SNRs $\leq 1$) tensors. From these results one can observe that amplification achieves the best performance for uniformly sized tensors, with the multiway Wiener filter achieving the second-best denoising performance. In the case of non-uniformly sized tensors, the Wiener filter and amplification achieve comparable results for starting SNRs of 1 and $-1$ dB, while amplification achieving better performance for starting SNRs of $-5$,$-10$, and $-20$ dB.
\begin{figure}[h!]
    \begin{subfigure}{0.5\textwidth}
    \includegraphics[width=\textwidth]{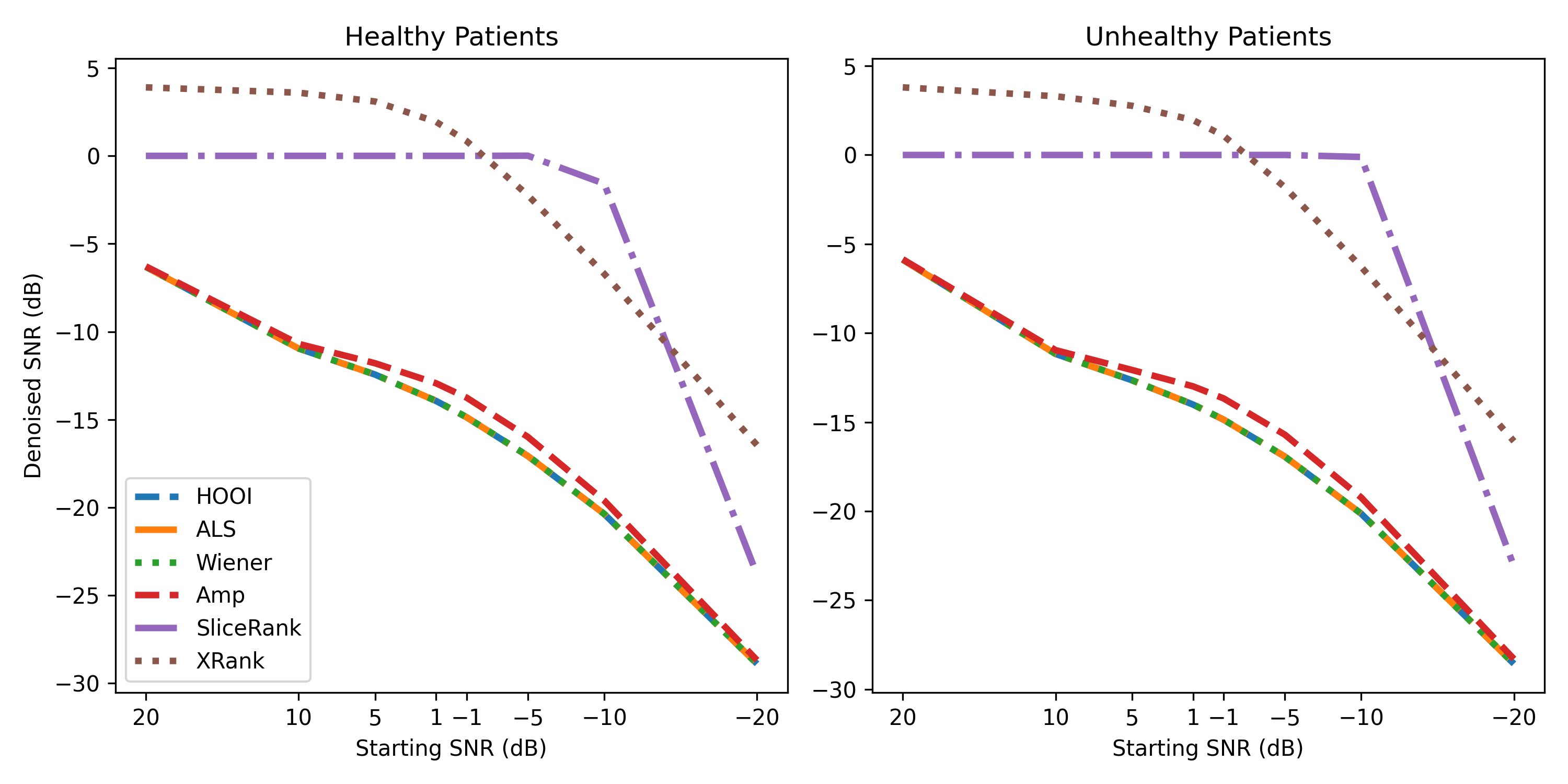}
    \caption{}
    \end{subfigure}
    \begin{subfigure}{0.5\textwidth}
    \includegraphics[width=\textwidth]{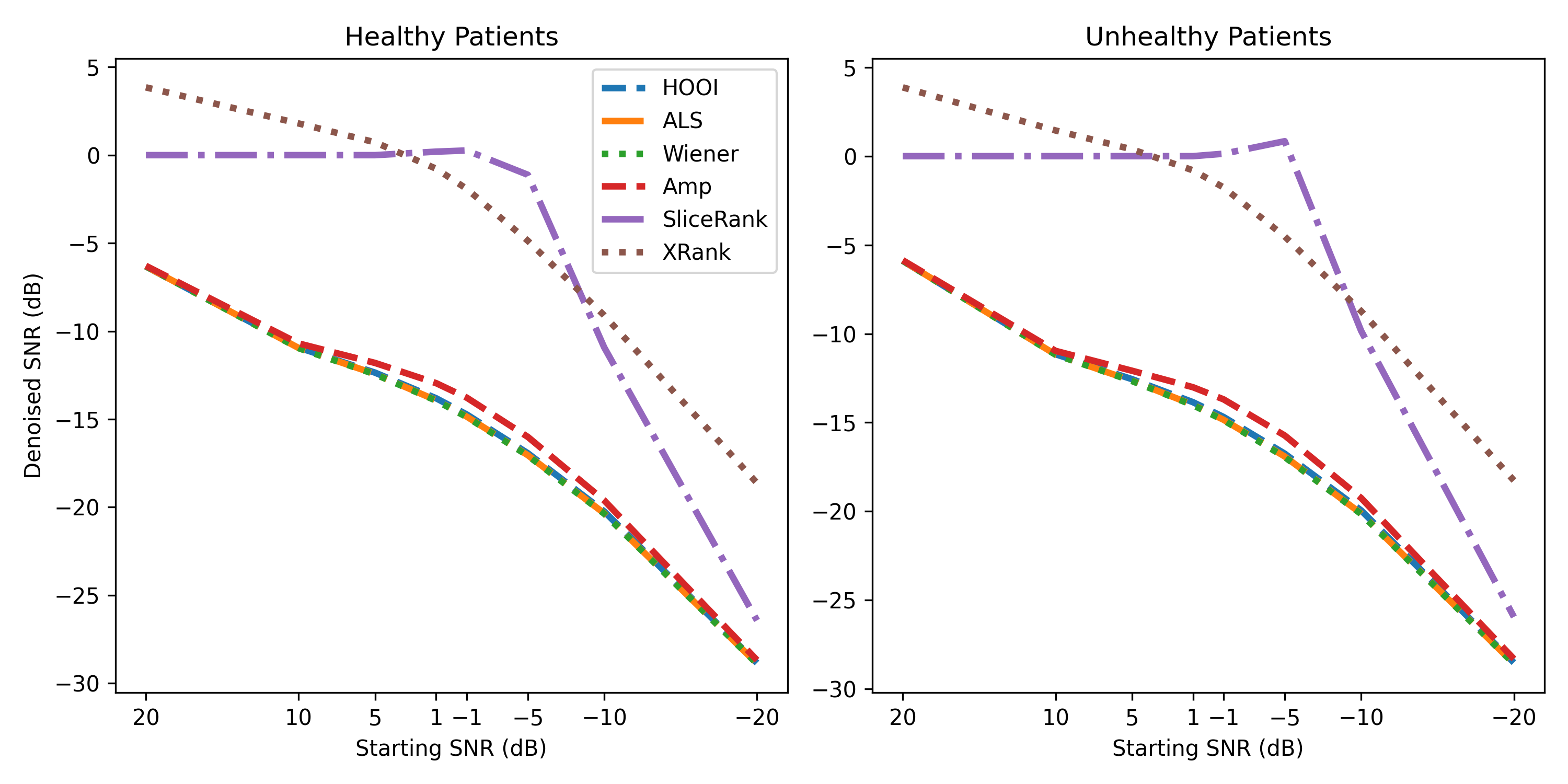}
    \caption{}
    \end{subfigure}
     \caption{Denoising performance on the PTB tensors formed from a) 90 second samples, and b) windowed samples.}
   \label{fig:ptb}
\end{figure}
\subsection{Results: Real Data - ECG Waveform Tensors}

Figure \ref{fig:ptb} shows the denoising performance on the tensors derived from the PTB dataset. For the tensors derived from 90 second samples of ECG signal (Figure \ref{fig:ptb} a), only the stable rank methods (XRank and SliceRank) were able to achieve any effective denoising, with XRank achieving the best denoising performance with a modest $\approx 4$ dB denoised SNR for tensors whose signals had SNR ratios of 20 dB \textit{prior to} tensor formation. This performance was maintained for tensors from signals with starting SNRs down to 5 dB, after which the denoising performance of XRank declines. The SliceRank method does not yield any tensor denoising until the starting signal SNR dropped below $-5$ dB, after which it too experiences a continued decline in denoising performance. All other methods - HOOI, ALS, and amplification - \textit{introduced} noise into the tensors across all starting signal SNRs. No appreciable difference was observed in tensors derived from the two patient cohorts (healthy and unhealthy). The denoising results for the tensors derived from windowed samples (Figure \ref{fig:ptb} b) are essentially the same as those derived from the 90 second samples, with the only exception being a slight increase in SliceRank's denoising performance for tensors corresponding to unhealthy patients with a starting signal SNR of $-5$ dB.

\FloatBarrier
\subsection{Discussion}

Overall alternating least squares (ALS) was the best performing method for denoising synthetic tensors across all tensor orders, sizes, ranks, and starting noise levels, with the multiway Wiener filter (MWF) also performing well across all parameters. Amplification-based denoising performed well for low ranked tensors as well as very noisy ($<0$ dB) tensors. The performance of amplification at low ranks and its decreased performance at higher ranks is to be expected, as the amplification maps correspond to approximations of the spectral norm, which only measures the highest singular value for a given tensor. Amplification-based denoising for higher rank tensors may be improved through the development of a decomposition method that can find successively smaller singular values and their corresponding rank 1 components, such as through a gradient-based descent optimization method.

For the tensors derived from physiological signals, only the XRank method had any appreciable denoising performance. Such a method may find applications as a preprocessing step in a machine learning pipeline that utilizes tensorial data, such as that used for the prediction of hemodynamic decomposition in \cite{hernandez2021multimodal}. One limitation of the amplification-based denoising method is that amplification requires determining order-specific amplification maps; currently only those for orders three and four have been computed. Other tested methods have no such restriction. However, many real-world data modalities, such as images (order 3) and video (order 4) can potentially be denoised using current amplification maps.\\

{\rowcolors{2}{cyan!5}{cyan!10}
\begin{table}[h]
    \centering
    \begin{tabular}{|c|c|c|}
    \hline
\textbf{Rank}	&\textbf{Low Noise ($\ge 5$ dB)}&\textbf{High Noise ($<5$ dB)}\\
\hline
Rank 1	& Amplification	& Amplification\\
Sparse (Low Rank, $\leq 5$) &	Wiener	& Amplification, Wiener\\
Dense (Rank $>5$)	&ALS, Wiener &	ALS, Wiener\\
\hline
    \end{tabular}
    \caption{Guidance for choosing a denoising algorithm for a given application based on tensor density and typical SNRs.}
    \label{tab:guidance}
\end{table}}

Based on the results presented above, Table \ref{tab:guidance} provides a useful summary for which denoising algorithm to choose for a given application, provided that some estimates of the typical tensor rank and noise level are known.  Given that tensor amplification can produce the best rank-1 approximation of a tensor \cite{tokcan2021algebraic}, amplification-based denoising should be used in the restricted case of rank 1 tensors.  For applications involving sparse data (ranks less than 5) and high SNRs, the multiway Wiener filter is a good choice. For low SNRS and low rank applications, amplification- and Wiener-based methods perform similarly. Finally, for denoising applications involving dense (higher rank) tensors, ALS is the preferred choice.

\FloatBarrier

\section{Conclusion}
In this work, we utilize the general framework of tensor denoising introduced \cite{derksen2018general} and previously developed approximations of the spectral norm \cite{tokcan2021algebraic} to devise three novel tensor denoising methods based on tensor amplification and two notions of tensor rank related to the $G$-stable rank \cite{derksen2022g} - stable slice rank and stable X-rank. The performance of these methods was compared to several standard decomposition-based denoising methods on synthetic tensors of various sizes, ranks, and noise levels, along with real-world tensors derived from electrocardiogram (ECG) signals. The experimental results show that in the low rank (sparse) context, tensor-based amplification provides comparable denoising performance in high signal-to-noise ratio (SNR) settings ($> 0$ dB) and superior performance in noisy ($<1$ dB) settings, while the stable $X$-rank method achieves superior denoising performance on the ECG signal data. Future work will seek to improve the performance of amplification-based methods for higher rank tensors.
\section*{Acknowledgment}
This work was partially supported by the National Science Foundation under Grant No. 1837985 and by the Department of Defense under Grant No. BA150235.
\appendices
\section{Order 4 Denoising Results}
\label{sec:order4results}
\begin{table*}
\centering
\caption{Mean (SD) SNR, in decibels, after tensor denoising across all parameters.}
\begin{tabular}{|c|cccccc|}
\hline
\textbf{Starting SNR} &          \textbf{HOOI} &            \textbf{ALS} &         \textbf{Wiener} &            \textbf{Amp} &     \textbf{SliceRank} &         \textbf{XRank} \\
\hline
 20 & 19.68 (3.70) & \textbf{33.04 (12.81)} & \textit{32.12 (15.50)} & 10.93 (15.73) & 18.96 (1.43) & 9.82 (4.10) \\
10 & 10.82 (1.35) & \textbf{24.54 (9.92)} & \textit{22.99 (11.98)} & 9.60 (12.65) & 13.31 (2.18) & 9.20 (3.60) \\
5 & 6.11 (0.88) & \textbf{20.04 (8.73)} & \textit{17.08 (10.24)} & 8.81 (11.16) & 8.29 (2.63) & 8.31 (3.10) \\
1 & 2.20 (0.82) & \textbf{16.21 (8.10)} & \textit{11.73 (9.21)} & 7.92 (10.06) & 3.62 (2.38) & 7.05 (2.67) \\
-1 & 0.22 (0.81) & \textbf{14.24 (7.85)} & \textit{8.90 (8.47)} & 7.36 (9.60) & 1.23 (2.12) & 6.22 (2.58) \\
-5 & -3.78 (0.80) & \textbf{10.14 (7.64)} & 3.68 (6.62) & \textit{6.06 (8.98)} & -3.44 (1.61) & 4.10 (2.77) \\
-10 & -8.79 (0.79) & \textbf{4.89 (7.52)} & -2.18 (4.03) & \textit{3.93 (8.78)} & -9.03 (1.09) & 0.69 (3.64) \\
-20 & -18.78 (0.81) & \textbf{-5.44 (7.21)} & \textit{-9.70 (6.52)} & -16.67 (3.64) & -19.64 (0.49) & -7.44 (5.15) \\
\hline
\multicolumn{7}{c}{(a) Uniformly sized tensors of order 4.}\\
\hline
\textbf{Starting SNR} &          \textbf{HOOI} &            \textbf{ALS} &         \textbf{Wiener} &            \textbf{Amp} &     \textbf{SliceRank} &         \textbf{XRank} \\
\hline
 20 & 25.83 (8.53) & \textbf{35.74 (14.42)} & \textit{32.02 (15.29)} & 14.20 (15.98) & 18.82 (1.44) & 11.16 (4.57) \\
10 & 17.69 (5.21) & \textbf{28.12 (11.06)} & \textit{24.55 (12.18)} & 12.83 (12.86) & 13.34 (2.14) & 10.41 (3.85) \\
5 & 13.47 (3.77) & \textbf{24.00 (9.59)} & \textit{18.98 (11.13)} & 11.83 (11.43) & 8.32 (2.49) & 9.31 (3.17) \\
1 & 10.00 (2.87) & \textbf{20.60 (8.58)} & \textit{13.77 (10.31)} & 10.14 (10.68) & 3.67 (2.21) & 7.90 (2.67) \\
-1 & 8.21 (2.55) & \textbf{18.86 (8.14)} & \textit{10.99 (9.52)} & 8.70 (10.62) & 1.29 (1.99) & 6.97 (2.49) \\
-5 & 4.49 (2.20) & \textbf{15.20 (7.47)} & \textit{6.25 (7.26)} & 4.99 (11.29) & -3.41 (1.49) & 4.63 (2.69) \\
-10 & -0.49 (2.06) & \textbf{10.30 (7.00)} & \textit{0.78 (4.80)} & 0.76 (12.49) & -8.95 (1.16) & 0.91 (3.77) \\
-20 & -10.64 (1.95) & \textbf{-0.10 (6.65)} & \textit{-6.69 (6.44)} & -18.46 (2.99) & -19.53 (0.67) & -7.47 (5.21) \\
\hline
\multicolumn{7}{c}{(b) Non-uniformly sized tensors of order 4.}
\end{tabular}
\label{tab:4overallsnr}
\end{table*}

\begin{figure*}
    \begin{subfigure}{0.5\textwidth}
    \includegraphics[width=\textwidth]{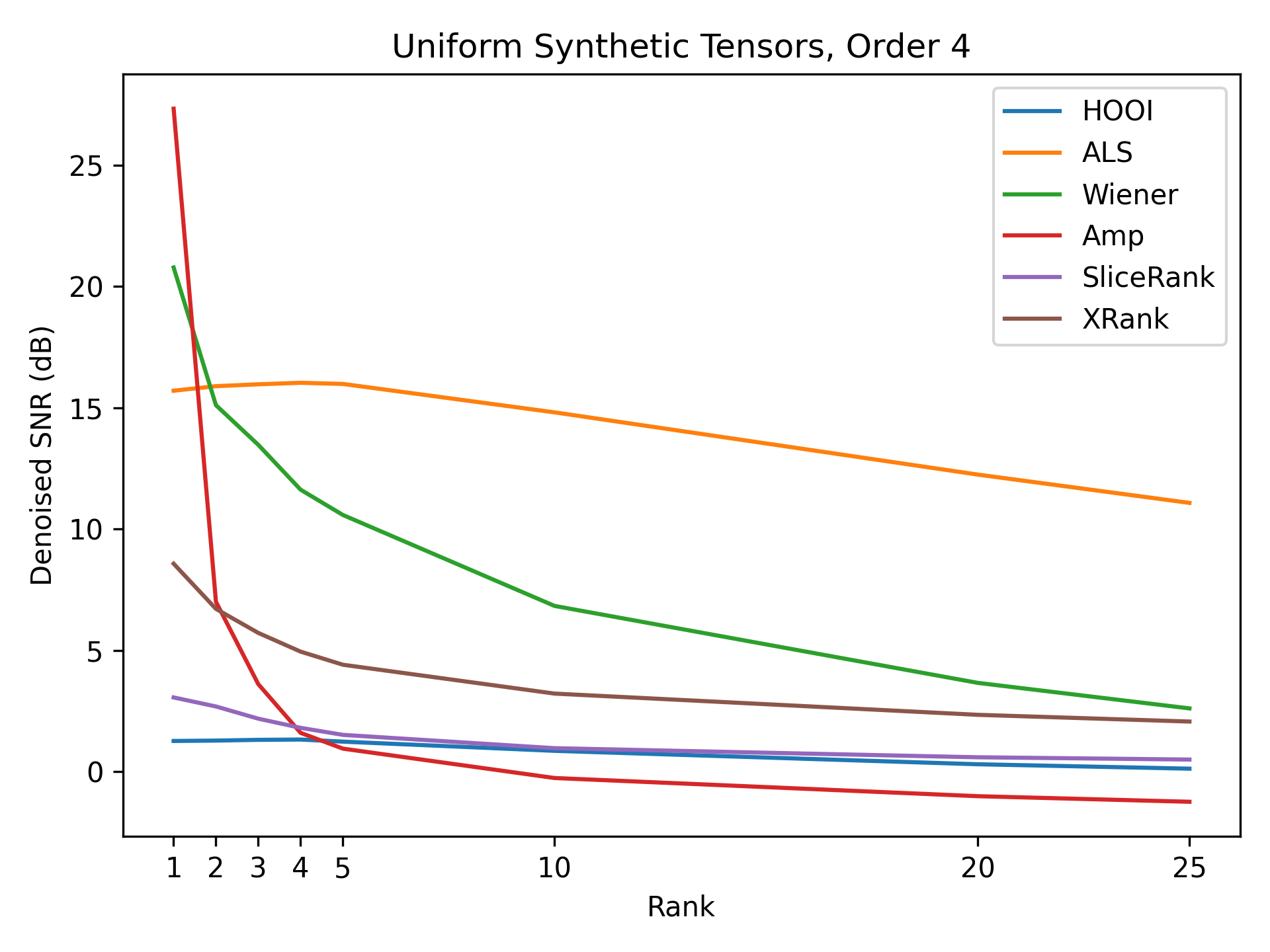}
    \caption{}
    \end{subfigure}
    \begin{subfigure}{0.5\textwidth}
    \includegraphics[width=\textwidth]{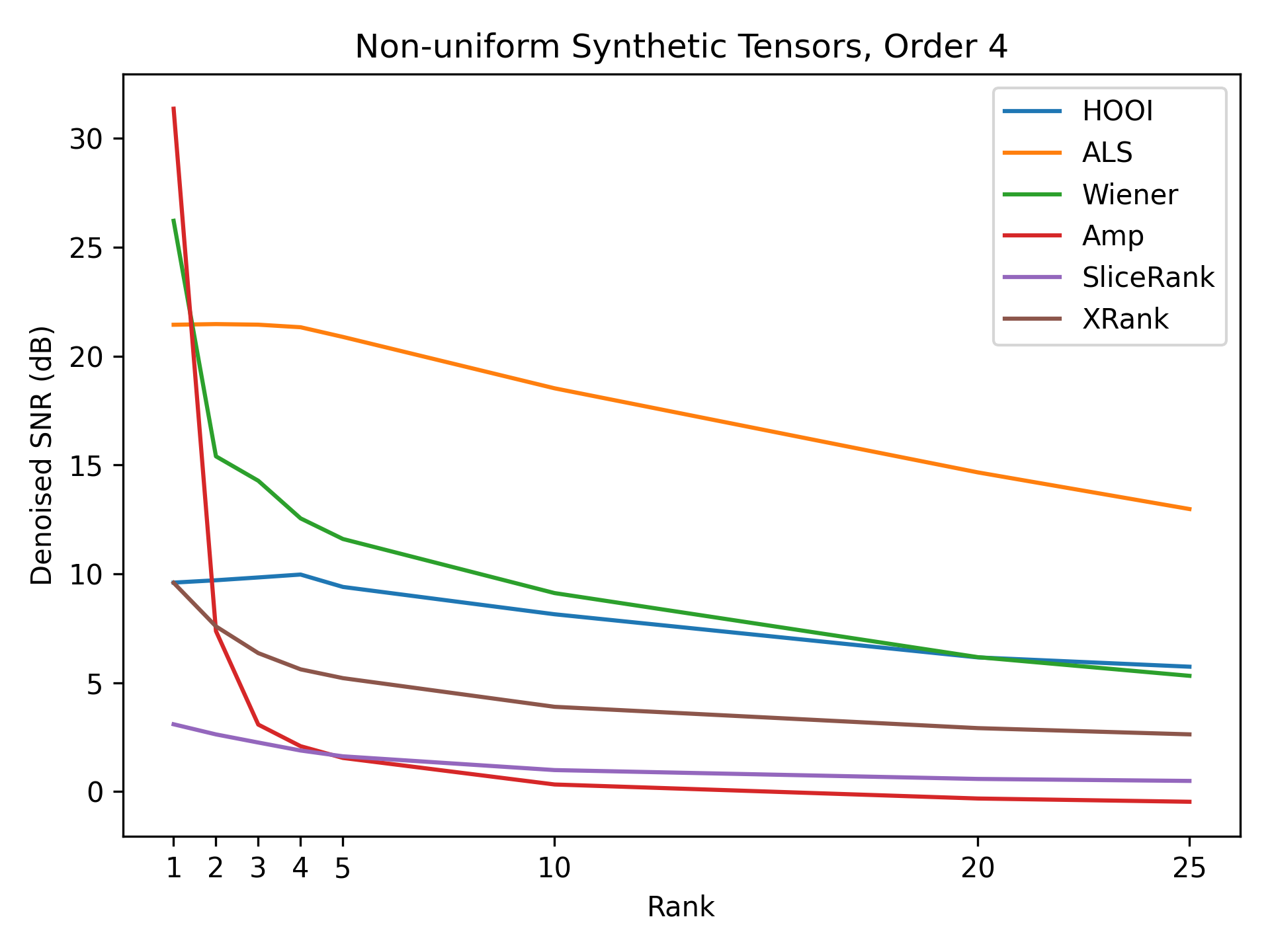}
    \caption{}
    \end{subfigure}
     \caption{Denoising performance with respect to tensor rank for fourth-order tensors.}
   \label{fig:4synthetic-ranks}
\end{figure*}

\begin{table*}
\centering
\caption{Mean (SD) denoised SNR, in decibels, for low rank and noisy tensors.}
\begin{tabular}{|c|cccccc|}
\hline
\textbf{Starting SNR} &          \textbf{HOOI} &            \textbf{ALS} &         \textbf{Wiener} &            \textbf{Amp} &     \textbf{SliceRank} &         \textbf{XRank} \\
\hline
1 & 1.59 (0.36) & 5.97 (2.93) & \textit{12.21 (4.05)} & \textbf{14.98 (7.72)} & 5.64 (2.29) & 8.89 (1.91) \\
-1 & -0.42 (0.35) & 3.92 (2.92) & \textit{10.31 (3.85)} & \textbf{13.66 (7.28)} & 3.33 (2.11) & 7.34 (2.01) \\
-5 & -4.43 (0.32) & -0.17 (2.91) & \textit{6.42 (3.78)} & \textbf{10.68 (6.90)} & -1.52 (1.70) & 3.77 (2.34) \\
-10 & -9.45 (0.30) & -5.24 (2.93) & \textit{1.90 (3.84)} & \textbf{6.16 (7.47)} & -7.49 (1.39) & -0.79 (2.88) \\
-20 & -19.45 (0.30) & -15.29 (2.91) & \textit{-8.86 (5.90)} & \textbf{-4.36 (7.84)} & -18.79 (1.03) & -10.31 (3.48) \\
\hline
\multicolumn{7}{c}{(a) Uniformly sized fourth-order tensors.}\\
\hline
\textbf{Starting SNR} &          \textbf{HOOI} &            \textbf{ALS} &         \textbf{Wiener} &            \textbf{Amp} &     \textbf{SliceRank} &         \textbf{XRank} \\
\hline
1 & 10.68 (2.08) & 22.45 (7.05) & \textbf{24.99 (11.39)} & \textit{23.47 (14.61)} & 5.82 (2.20) & 10.97 (2.55) \\
-1 & 8.67 (2.08) & 20.36 (7.02) & \textit{21.36 (11.26)} & \textbf{22.36 (13.82)} & 3.43 (2.08) & 9.71 (2.59) \\
-5 & 4.65 (2.07) & \textit{16.19 (6.95)} & 14.20 (9.11) & \textbf{20.10 (12.42)} & -1.59 (1.63) & 6.94 (2.98) \\
-10 & -0.38 (2.05) & \textit{11.03 (6.88)} & 4.95 (5.79) & \textbf{17.07 (10.93)} & -7.48 (1.42) & 3.01 (4.21) \\
-20 & -10.56 (1.98) & \textbf{0.71 (6.97)} & \textit{-6.40 (5.90) }& -15.23 (4.45) & -18.66 (0.87) & -5.51 (5.53) \\
\hline
\multicolumn{7}{c}{(b) Non-uniformly sized fourth-order tensors.}
\end{tabular}
\label{tab:4lowranklownsr}
\end{table*}
\pagebreak
\FloatBarrier
\bibliographystyle{IEEEtran}
\bibliography{tensor_denoising.bib}

%





\end{document}